%% file: main.tex
\RequirePackage{amsmath,amssymb}
\documentclass[10pt,letterpaper]{sharpa}
\input{preamble}
\title{WiTac: Wrist-Informed Tactile Representations\\for Contact-Rich Manipulation}
\input{authors}
\abstract{\input{sections/0_abstract}}
\metadata[Correspondence]{\sffamily\href{mailto:zhijied@sjtu.edu.cn}{zhijied@sjtu.edu.cn}}
\hypersetup{pdftitle={WiTac: Wrist-Informed Tactile Representations for Contact-Rich Manipulation}}
\begin{document}
\maketitle
\input{figures/teaser}
\input{sections/1_introduction}
\input{sections/2_related_work}
\input{sections/3_method}
\input{sections/4_experiments}

\input{sections/5_conclusion}

\label{maintext:end}
\FloatBarrier
\setlength{\bibsep}{2pt}
\bibliographystyle{plainnat}
\bibliography{references}
\clearpage
\pdfbookmark[0]{Appendix}{witac.appendix}
\section*{Appendix}
\vspace{-4pt}
\input{appendix}
\end{document}

%% file: preamble.tex
\usepackage{lmodern}

\usepackage{xspace}
\newcommand{\witacnw}{\mbox{WiTac-NW}\xspace}
\usepackage[shortlabels,inline]{enumitem}
\usepackage{adjustbox}
\usepackage{wrapfig,needspace}

\input{figures/figure_styles}
\setcitestyle{numbers,square,comma}
\renewcommand{\titlelogo}{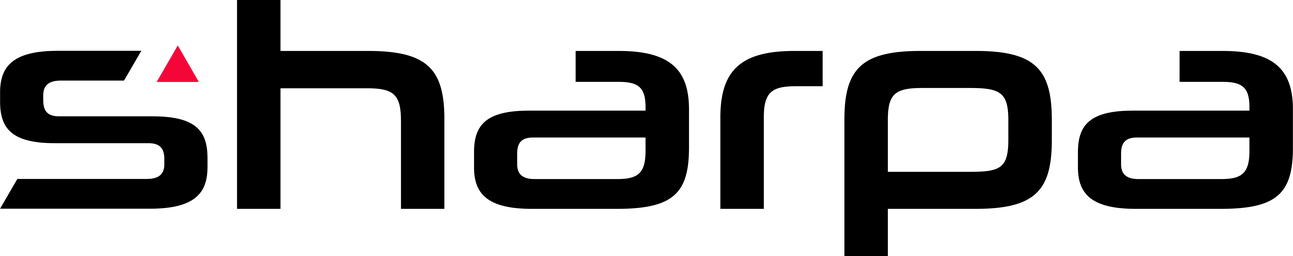}
\renewcommand{\titlelogowidth}{2.8cm}
\renewcommand{\titlelogoraise}{-16.7pt}

%% file: figures/figure_styles.tex
\usepackage{tikz}
\usetikzlibrary{arrows.meta,positioning,calc,fit,backgrounds}
\definecolor{twTouch}{RGB}{0,133,108}
\definecolor{twVision}{RGB}{55,109,159}
\definecolor{twJoint}{RGB}{104,72,161}
\definecolor{twTrain}{RGB}{181,111,23}
\definecolor{twGray}{RGB}{100,108,118}
\tikzset{
 tw/base/.style={rounded corners=2pt,line width=.65pt,align=center,
  font=\sffamily\fontsize{7}{8.5}\selectfont,inner sep=4pt,minimum height=.66cm},
 tw/input/.style={tw/base,draw=twGray!65,fill=twGray!4},
 tw/touch/.style={tw/base,draw=twTouch,fill=twTouch!8},
 tw/vision/.style={tw/base,draw=twVision,fill=twVision!7},
 tw/joint/.style={tw/base,draw=twJoint,fill=twJoint!9},
 tw/train/.style={tw/base,draw=twTrain,fill=twTrain!7},
 tw/policy/.style={tw/base,draw=twGray,fill=twGray!5},
 tw/flow/.style={-{Latex[length=1.6mm]},draw=twGray,line width=.7pt},
 tw/tflow/.style={tw/flow,draw=twTouch},
 tw/vflow/.style={tw/flow,draw=twVision},
 tw/jflow/.style={tw/flow,draw=twJoint},
 tw/lflow/.style={tw/flow,draw=twTrain},
 tw/note/.style={font=\sffamily\fontsize{7}{8}\selectfont,text=twGray,align=center},
 tw/title/.style={font=\sffamily\fontsize{8}{9.5}\selectfont\bfseries,anchor=west},
}

%% file: authors.tex
\renewcommand{\authorformat}[2][]{%
  \mbox{{\fontsize{11}{14}\selectfont\sffamily\bfseries #2}$^{#1}$}}
\renewcommand{\affiliationformat}[2][]{%
  \mbox{{\fontsize{9}{12}\selectfont\sffamily $^{#1}$#2}}}
\renewcommand{\authorlist}{%
  \authorformat[1,2,*]{Haotian He},
  \authorformat[1,3,*]{Yiyang Chen},
  \authorformat[1,3]{Congsheng Xu},
  \authorformat[1,5]{Wenchao Yue}\\[3pt]
  \authorformat[1,3]{Mingzhe Ye},
  \authorformat[1,4]{Shuai Chen},
  \authorformat[1,3]{Yi Yang},
  \authorformat[1,5]{Yaping Li}\\[3pt]
  \authorformat[1]{Haonan Wang},
  \authorformat[2]{Weinan E},
  \authorformat[3]{Wenzhao Lian},
  \authorformat[3]{Zhijie Deng}}

\affiliation[1]{Sharpa Robotics}
\affiliation[2]{PKU}
\affiliation[3]{SJTU}
\affiliation[4]{ShanghaiTech}
\affiliation[5]{CUHK}
\contribution[*]{Equal contribution}
\hypersetup{pdfauthor={Haotian He, Yiyang Chen, Congsheng Xu, Wenchao Yue, Mingzhe Ye, Shuai Chen, Yi Yang, Yaping Li, Haonan Wang, Weinan E, Wenzhao Lian, Zhijie Deng}}

%% file: sections/0_abstract.tex
Similar tactile contacts can call for different corrections depending on the configuration of the hand, object and target.
Rather than supplying touch and wrist vision separately, fusing them before policy injection better grounds tactile features in visual context, yielding more accurate contact representations.
We introduce Wrist-Informed Tactile Representations (WiTac), a pretrained encoder that learns such representations before action generation.
Pretrained on nearly 300 hours of SharpaDex data, WiTac deeply fuses tactile signals with hand motion and wrist views into a comprehensive contact representation.
Conditioning downstream VLA policies (GR00T N1.7 and $\pi_{0.5}$) on WiTac substantially improves task success over vision-only base policies across four contact-rich manipulation tasks (+28.8\% and +15.0\% on average), and further outperforms a wrist-free tactile encoder by 18.8\% and 7.5\%, respectively.
Offline analyses further show that incorporating wrist vision yields more accurate contact descriptors, stronger joint task--contact discrimination, greater feature stability under tactile noise, and clearer action-direction readout from similar contacts.
Project page: \url{https://wrist-informed-tactile-representations.github.io/}.

%% file: figures/teaser.tex
\begin{figure}[t]
\centering
\includegraphics[width=\linewidth]{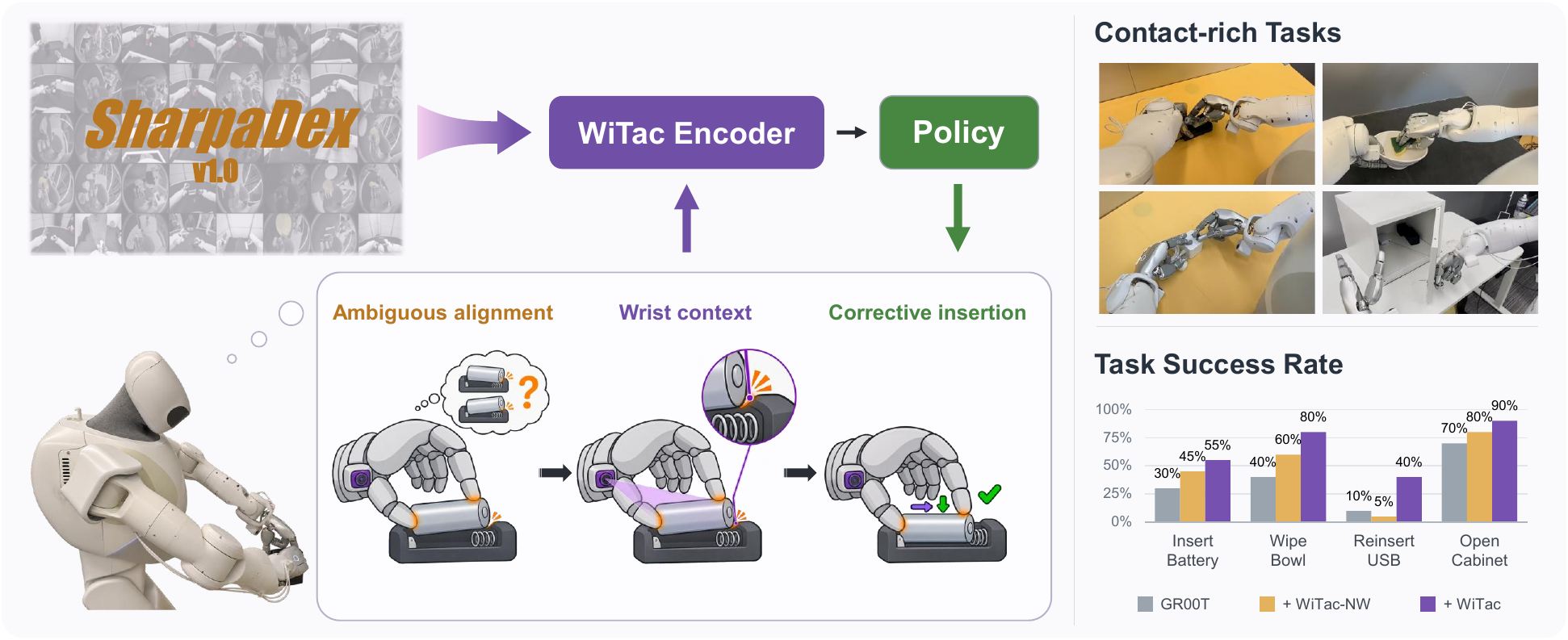}
\captionsetup{hypcap=false}
\caption{\textbf{Wrist context complements fingertip contact.}
Left: WiTac is pretrained on SharpaDex v1.0~\cite{sharpadex_v1_2026} and conditions a downstream policy. The battery-insertion cartoon illustrates how wrist context helps resolve alignment ambiguity during contact. Right: evaluation tasks and success rates.}
\label{fig:concept}
\end{figure}

%% file: sections/1_introduction.tex
\section{Introduction}
\label{sec:intro}

Contact-rich manipulation requires actions that respond jointly to local touch and surrounding task geometry.
When inserting a battery into a slot, similar finger loading can accompany different battery--slot alignments and call for distinct corrective motions.
Local tactile deformation alone leaves this spatial configuration ambiguous, whereas close-range wrist cameras observe the hand, grasped object, and nearby target geometry, providing the spatial reference needed to interpret contact during manipulation.

Generalist vision-language-action (VLA) policies such as GR00T~\cite{gr00tn1_2025} and $\pi_{0.5}$~\cite{pi05_2025} already receive camera images in their vision-language streams, yet they lack pre-fusion with tactile signals to ground grasp- and contact-related states.
Recent visuotactile designs incorporate touch into policies by supplying standalone tactile features alongside visual streams~\cite{sparsh2024,simpletouch2026}, modulating visual features with tactile signals~\cite{tacfilm2026}, or encoding discrete residuals from visually predicted touch~\cite{restacvla2026}.
However, separate tactile inputs leave local touch ungrounded in the surrounding visual geometry, while visual modulation and residual coding do not directly form continuous contact representations that preserve detailed touch within its wrist visual context.

To address this limitation, we introduce Wrist-Informed Tactile Representations (WiTac), a pretrained encoder that fuses touch with wrist vision before policy injection to produce grounded visuotactile contact tokens (\cref{fig:concept}).
Pretrained on nearly 300 hours of synchronized dexterous manipulation data from SharpaDex v1.0~\cite{sharpadex_v1_2026}, WiTac summarizes each finger's tactile history with robot state and wrist-relative fingertip kinematics, then queries wrist patch features through cross-attention to form finger-level contact tokens.
Pretraining jointly supervises contact-state and history reconstruction, masked wrist-feature recovery, and action-conditioned wrench-response prediction, encouraging the tokens to retain physical contact detail as they incorporate visual context.
During policy learning, the frozen WiTac encoder conditions VLA policies through a lightweight gated cross-attention interface alongside their native visual, language and state inputs.
Across four real-world contact-rich tasks, WiTac consistently boosts closed-loop success on both GR00T N1.7 and $\pi_{0.5}$ over vision-only base policies (+28.8\% and +15.0\% on average) and its wrist-free variant \witacnw (+18.8\% and +7.5\%), while offline analyses confirm improved contact reconstruction, joint task--contact discrimination, noise stability, and action-direction readout.

Our contributions are:
\begin{itemize}
    \item We propose WiTac, a pretrained contact encoder that fuses tactile histories, hand kinematics, and wrist observations into finger-level contact tokens before action generation, through joint physical and visual supervision.
    \item We demonstrate the utility of frozen WiTac representations across two VLA backbones and four real-world contact-rich tasks, achieving higher success than vision-only policies and wrist-free tactile encoding on every task.
    \item Through representation analyses, we show that WiTac provides more accurate contact descriptors, stronger joint task--contact discrimination, lower feature drift under tactile noise, and more accurate action-direction readout than wrist-free encoding.
\end{itemize}

%% file: sections/2_related_work.tex
\section{Related Work}
\label{sec:related}

\paragraph{Tactile and visuotactile representations.}
Pretrained tactile encoders provide reusable features for contact perception and manipulation.
Sparsh~\cite{sparsh2024} learns touch representations through self-supervised objectives on tactile images, while T3~\cite{t3_2025} combines sensor-specific encoders with a shared Transformer across sensors and tasks.
AnyTouch~\cite{anytouch2025} unifies static tactile images and dynamic tactile videos across multiple visuo-tactile sensors through masked modeling and cross-sensor alignment, and Sparsh-X~\cite{sparshx2025} extends representation learning to image, audio, motion, and pressure signals from a multisensory tactile sensor.
Physical supervision offers another route: UniVTAC~\cite{univtac2026} reconstructs tactile appearance and predicts deformation, marker locations, and relative object pose from simulated tactile observations.
External vision and language can also shape tactile representations: early work uses self-supervision to learn compact visuotactile features for contact-rich policy learning~\cite{lee2018visionsense}, VITaL~\cite{vital2024} aligns visual and tactile embeddings through temporal contrastive pretraining, OmniVTLA~\cite{omnivtla2025} aligns touch with visual and linguistic concepts in a dual-encoder architecture, and VT-MUSE~\cite{vtmuse2026} pretrains sequential visuotactile latents through cross-modal alignment and masked sensory reconstruction.
% WiTac instead forms wrist-conditioned contact tokens directly at deployment: per-finger tactile history, wrist features from a frozen DINOv3~\cite{dinov3_2025} backbone, and robot state are fused within one encoder under multi-objective contact, motion, visual, and wrench supervision before conditioning action generation.

\paragraph{Visuotactile policies for parallel-jaw grippers.}
Vision-language-action (VLA) models such as GR00T~\cite{gr00tn1_2025} and $\pi_{0.5}$~\cite{pi05_2025} provide generalist visual, language, and motor capabilities that recent gripper-based systems extend with touch through policy conditioning or predictive world modeling.
For reactive control, RDP~\cite{rdp2025} pairs a slow latent diffusion policy with high-frequency closed-loop tactile feedback.
Within VLA policies, SimpleTouch~\cite{simpletouch2026} passes tokens from a frozen pretrained tactile encoder to a tactile expert trained with action supervision and future-latent prediction, TacFiLM~\cite{tacfilm2026} maps pretrained tactile features to feature-wise scale and shift parameters that modulate the visual encoder, and ResTacVLA~\cite{restacvla2026} encodes the quantized residual between wrist-predicted and observed touch with uncertainty gating.
To anticipate contact evolution, DreamTacVLA~\cite{dreamtacvla2025} aligns multi-scale views with touch and refines draft actions using future tactile states predicted by a world model, while OmniVTA~\cite{omnivta2026} couples a two-stream visuo-tactile world model and contact-aware fusion policy with a 60\,Hz reflexive controller.
World-action models further unify dynamics prediction and control: VTAM~\cite{vtam2026} treats optical tactile images as an additional view of a pretrained video backbone with a flow-derived force proxy, Dream-Tac~\cite{dreamtac2026} jointly predicts future visual observations, tactile dynamics, and actions using a contact-aware attention bias, and FAWAM~\cite{fawam2026} and ForceWAM~\cite{forcewam2026} predict future wrenches alongside actions.

\paragraph{Scaling tactile manipulation to dexterous hands.}
Moving from parallel-jaw grippers to multi-fingered dexterous hands introduces distributed contacts across fingers, higher-dimensional kinematics, and heavier visual self-occlusion.
To bridge embodiments, FTP-1~\cite{ftp1_2026} maps heterogeneous tactile signals into morphology-aware tokens processed by a shared tactile Transformer expert, and ME-Dex 1.0~\cite{medex2026} aligns gripper and dexterous-hand tactile inputs onto a canonical hand template within a three-expert world-action model.
For multi-fingered hands specifically, SaTA~\cite{sata2025} anchors tactile features to wrist-frame fingertip poses, ViTacFormer~\cite{vitacformer2025} fuses vision and touch via cross-attention with an autoregressive tactile-prediction head, and STAR~\cite{star2026} jointly predicts masked wrist-image and tactile patches during pretraining before passing separate tactile tokens to a VLA policy.
At the policy and world-model level, T-Rex~\cite{trex2026} combines a temporal tactile VQ-VAE with a variable-rate Mixture-of-Transformers to exploit high-frequency tactile feedback, DeCAL~\cite{decal2026} pairs contact-gated fusion with latent visuo-tactile co-imagination, and DexTacWAM~\cite{dextacwam2026} compresses per-finger touch into a video diffusion world model to jointly predict visual and contact dynamics.
Rather than training a generative video-tactile world model during policy learning or supplying deployment tactile tokens encoded without wrist views, WiTac pretrains a frozen wrist-informed contact encoder that fuses ten-finger tactile histories, kinematics, and wrist imagery before conditioning VLA policies through lightweight gated cross-attention.

%% file: sections/3_method.tex
\section{Wrist-Conditioned Contact Representation}
\label{sec:method}
\input{figures/overview_float}

WiTac fuses fingertip touch, hand kinematics, and wrist views into finger-level contact tokens before action generation. We pretrain the encoder via self-supervision, then freeze it to condition a VLA policy through gated cross-attention. \Cref{fig:method}(a--c) overviews contact encoding, representation pretraining, and policy injection.

\subsection{Wrist-Informed Contact Encoding}
\label{sec:observations}
\label{sec:encoding}

Local fingertip deformation alone can be ambiguous without the surrounding hand--object configuration. Let $T_t$ denote the $240\times240$ scalar deformation maps of all ten fingers at time $t$, and $\mathcal T_t$ their eight-frame history including the current frame. WiTac receives $\mathcal T_t$, current RGB images $I_t$ from both wrist cameras, and the robot state $s_t$, while forward kinematics provides each finger's pose $p_t$ (3D position $x_t$ and orientation) in its respective wrist frame. From these inputs, the encoder produces one contact token per finger (\cref{fig:method}(a)):
\begin{equation}
 C_t = E_\theta(\mathcal{T}_t,I_t,s_t,p_t)
 \in \mathbb{R}^{10\times768}.
 \label{eq:contact}
\end{equation}

To encode tactile dynamics, a per-frame spatial backbone initialized from SharpaWave Deform Encoder~\cite{sharpawave_deform_encoder2026} maps each finger's deformation map to a 512-dimensional feature, and a causal temporal encoder compresses the eight-frame sequence into per-finger history features $H_t\in\mathbb{R}^{10\times512}$. Each finger feature is projected to the 768-dimensional fusion width and spatially anchored on Fourier-encoded wrist-relative fingertip poses $p_t$ via feature-wise linear modulation (FiLM), following SaTA~\cite{sata2025}. Adding a learned projection of the current robot state $s_t$ forms the state-conditioned tactile tokens entering cross-modal fusion.

In parallel, a frozen DINOv3~\cite{dinov3_2025} ViT-B/16 backbone extracts patch features $V_t$ concatenated across both wrist views, with camera-identity and 2D patch-coordinate embeddings specifying each patch's view and spatial location. Two Transformer blocks fuse the modalities: in each block, the tactile tokens first interact through finger self-attention, then query the wrist features $V_t$ as keys and values through cross-attention, followed by a feed-forward network. Final layer normalization yields $C_t$, while validity masks handle unavailable observations. Each contact token thus grounds a finger's tactile history in close-range wrist visual context before policy conditioning.

\subsection{Pretraining Contact Representations}
\label{sec:pretraining}

We pretrain WiTac on synchronized tactile, wrist-image, robot-state, and fingertip force--torque recordings from SharpaDex v1.0~\cite{sharpadex_v1_2026}. Five self-supervised objectives in three complementary groups (\cref{fig:method}(b)) supervise $C_t$ so that wrist fusion preserves local contact detail while incorporating visual context and action-dependent physical responses. All prediction heads read the shared token set $C_t$ during pretraining and are removed for policy inference. Finger-indexed outputs follow the ten-finger ordering of $C_t$, with hats denoting predictions.

\paragraph{Contact state and history reconstruction.}
Cross-attending to dense visual features risks washing out fine tactile signals. Following the sensory self-supervision rationale of Sparsh~\cite{sparsh2024}, three heads reconstruct current contact descriptors alongside recent deformation and fingertip-motion histories:
\begin{equation}
\begin{aligned}
 \widehat d_t &= D_{\mathrm{desc}}(C_t), &&\\
 \widehat{\Delta T}_{t,k} &= D_{\mathrm{def}}(C_t)_k,
 &\Delta T_{t,k} &= T_t-T_{t-k},\\
 \widehat{\Delta x}_{t,k} &= D_{\mathrm{mot}}(C_t)_k,
 &\Delta x_{t,k} &= x_t-x_{t-k}.
\end{aligned}
\label{eq:history_targets}
\end{equation}
Here $d_t$ contains contact area, deformation intensity, and 2D center for each finger, while $k$ indexes past-frame offsets from separate offset lists for deformation and motion. Because $x_t$ and $x_{t-k}$ are each expressed in their respective wrist frames at those timestamps, $\Delta x_{t,k}$ measures wrist-relative fingertip motion independent of global arm movement. Reconstructing these multi-step histories requires $C_t$ to retain recent contact changes beyond the current snapshot.

\paragraph{Masked visual recovery.}
Conversely, tactile objectives alone could let the fusion blocks ignore wrist imagery. Following feature-space masked prediction~\cite{ijepa2023}, we extract visible wrist features $V_t^{\mathrm{vis}}$ from a separate DINOv3 pass with native patch masking and decode masked patch features through $C_t$:
\begin{equation}
 \widehat V_t=D_{\mathrm{vis}}(C_t,V_t^{\mathrm{vis}}).
 \label{eq:visual_target}
\end{equation}
The loss compares $\widehat V_t$ with the detached full-image features $V_t$ only at masked patch positions. Because complete wrist features $V_t$ enter the fusion encoder, this objective trains conditional feature recovery rather than occluded-region inpainting. Routing information from 640 wrist patches through only ten finger tokens creates a bottleneck that compels $C_t$ to carry wrist visual context.

\paragraph{Wrench response prediction.}
To capture how the current contact configuration physically responds to upcoming actions~\cite{lee2018visionsense}, we train an action-conditioned head to predict fingertip wrench changes over a future horizon $h$:
\begin{equation}
 \widehat{\Delta w}_{t,h}=D_{\mathrm{wr}}(C_t,A_{t,h}),
 \qquad \Delta w_{t,h}=w_{t+h}-w_t.
 \label{eq:wrench_target}
\end{equation}
Here $w_t$ collects each finger's six-dimensional force--torque vector, and $A_{t,h}$ is the future action prefix up to $t+h$. Future actions enter only this training-time decoder, whereas $H_t$ and $C_t$ depend solely on causal observations available at time $t$.

We minimize the weighted objective
\begin{equation}
 \mathcal L_{\mathrm{pre}}=\sum_{j=1}^{5}\lambda_j\mathcal L_j,
 \label{eq:pretrain}
\end{equation}
with validity masks for missing entries and balanced weighting between changing and nearly static regions in deformation reconstruction. Gradients through $C_t$ jointly optimize the spatial and temporal tactile encoders together with the fusion blocks, while DINOv3 remains frozen. Temporal offsets, horizons, and objective weights are listed in \hyperref[sec:supp_pretraining]{Appendix~\ref*{sec:supp_pretraining}} (\cref{tab:pretraining_targets}), and pretraining uses no task-success labels.

\subsection{Gated Contact Injection into Action Policies}
\label{sec:injection}

For policy learning, we freeze WiTac and condition the action expert of a VLA policy through a shared gated cross-attention interface, while preserving the policy's native visual, language, and state inputs (\cref{fig:method}(c)). A learned projection $P$ maps $C_t$ to the hidden dimension of the action expert, and selected action blocks query the projected contact tokens as keys and values:
\begin{equation}
 X_l' = X_l + g_l(C_t)\,
       \operatorname{Cross\text{-}Attn}_l(X_l,P(C_t)),
 \label{eq:injection}
\end{equation}
where $X_l$ denotes the action features at block $l$. A shared gate predictor averages valid tokens in $C_t$, applies layer normalization and an MLP, and passes the outputs through a rescaled sigmoid function in $[0.05,1]$ to produce one scalar gate $g_l(C_t)$ per injected block. This bounded gate keeps a small nonzero floor when valid touch is present while scaling up the contact residual during physical interaction. Because the interface modifies only selected action blocks, a pretrained WiTac encoder can condition different VLA backbones through policy-specific projections and insertion layers.

We jointly optimize action prediction and contact-gate supervision:
\begin{equation}
\begin{aligned}
 \mathcal L_{\mathrm{policy}} &= \mathcal L_{\mathrm{action}}
                         +\lambda_{\mathrm{gate}}\mathcal L_{\mathrm{gate}},\\
 \mathcal L_{\mathrm{gate}} &= \operatorname{mean}_{t,l}
       \operatorname{BCE}\!\left(g_l(C_t),y_t\right),
\end{aligned}
\label{eq:policy_loss}
\end{equation}
where $\mathcal L_{\mathrm{action}}$ is the base policy's flow-matching loss and $\mathcal L_{\mathrm{gate}}$ averages binary cross-entropy over injected blocks and labeled timesteps. The contact proxy $y_t\in\{0,1\}$ is computed from the active-pixel fraction on the hands in $T_t$, encouraging stronger conditioning during physical contact. Throughout policy training, WiTac remains frozen while the contact interface and action expert are optimized jointly.

%% file: figures/overview_float.tex
\begin{figure}[t]
\centering
\includegraphics[width=\linewidth]{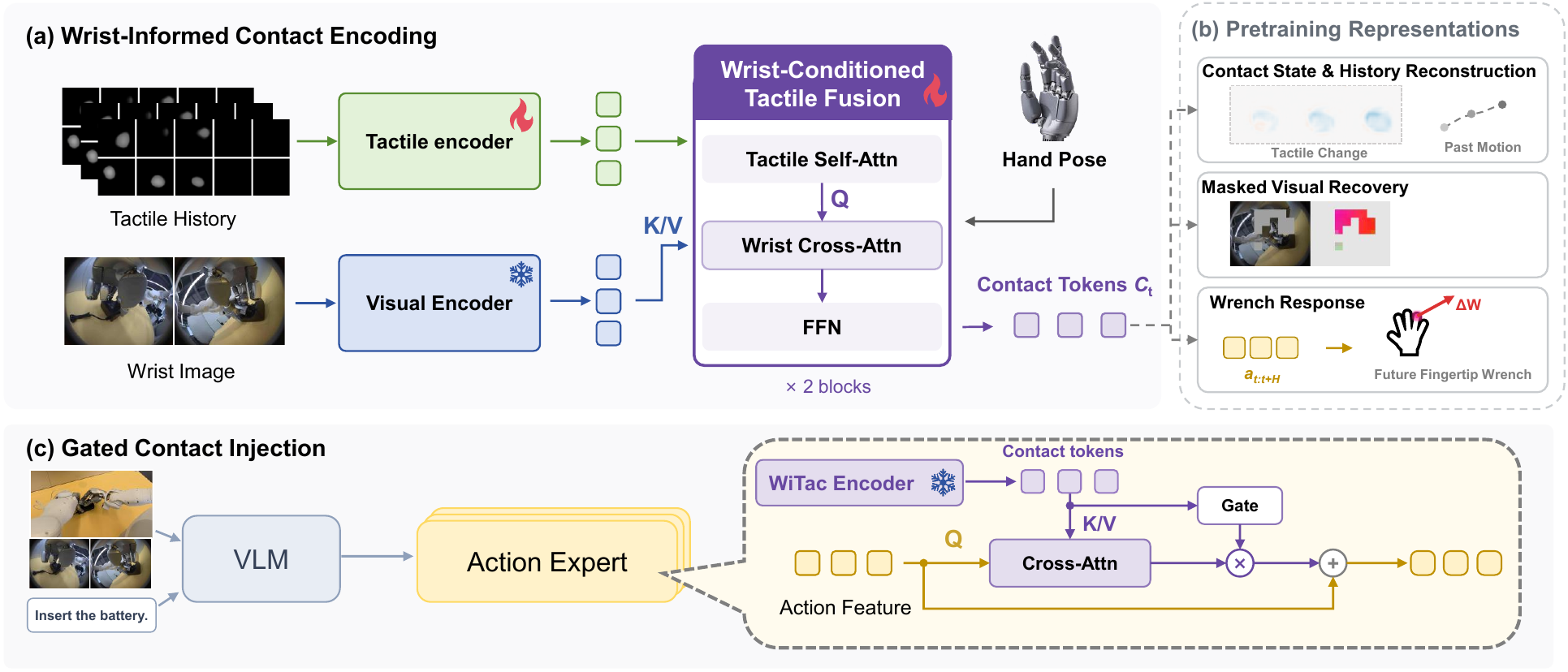}
\caption{\textbf{WiTac learns contact tokens before policy injection.}
(a) State-conditioned tactile histories query wrist features to form contact tokens $C_t$.
(b) Pretraining-only heads supervise contact state, deformation and fingertip-motion histories, conditional visual recovery, and action-conditioned wrench responses.
(c) Tokens from the frozen encoder condition the action expert through gated cross-attention.}
\label{fig:method}
\end{figure}

%% file: sections/4_experiments.tex
\section{Experiments}
\label{sec:exp}
Our experiments address three questions:
\begin{enumerate}
 \item[\textbf{Q1.}] Does WiTac improve real-world contact-rich manipulation over base policies and tactile encoders without wrist fusion?
 \item[\textbf{Q2.}] How does wrist conditioning affect contact reconstruction, context and action-direction readout, and sensitivity to tactile noise?
 \item[\textbf{Q3.}] How do representation pretraining, visual supervision, and pre-injection fusion contribute to task success?
\end{enumerate}

\subsection{Experimental Setup}
\label{sec:exp_setup}
\paragraph{Robot and tasks.}
We evaluate on a Sharpa North robot with two Sharpa Wave dexterous hands. Its 65-dimensional joint-command action space comprises 7 joints per arm, 22 per hand, 2 head/neck joints, and 5 lower-body joints. We evaluate four bimanual contact-rich tasks: battery insertion (seating a battery into a slot), bowl wiping (scrubbing the inner bowl wall with a sponge), USB reinsertion (extracting and reinserting a connector on a power strip), and cabinet opening (turning a key and pulling open a door), with detailed task definitions in \hyperref[sec:supp_setup]{Appendix~\ref*{sec:supp_setup}}. None of these four tasks are included in the SharpaDex v1.0 pretraining dataset; for each task, we collect 200 demonstrations through exoskeleton teleoperation for policy training.

\paragraph{Baselines.}
We integrate WiTac into GR00T N1.7~\cite{gr00tn1_2025} (GR00T hereafter) and $\pi_{0.5}$~\cite{pi05_2025}. For each backbone, we compare against the base policy and its wrist-free variant \witacnw, which removes only wrist cross-attention and masked visual recovery while keeping everything else unchanged. On GR00T, we additionally pretrain the visuotactile encoder of VT-MUSE~\cite{vtmuse2026} on SharpaDex v1.0 and inject it into GR00T (VT-MUSE joint repr.). We also compare with two standalone tactile policies: FTP-1~\cite{ftp1_2026}, which unifies heterogeneous sensor inputs through morphology-aware tactile tokens and a shared Transformer expert, and T-Rex~\cite{trex2026}, which couples a temporal tactile encoder with a variable-rate Mixture-of-Transformers architecture to incorporate high-frequency touch into action generation.
% All compared policies, including external baselines, are fine-tuned on the same 200 demonstrations per task, whereas only our contact encoders use SharpaDex v1.0 pretraining.

\paragraph{Evaluation protocols and metrics.}
Each method is evaluated over 20 trials per task. We additionally measure contact-related behaviors and analyze frozen representations through contact-reconstruction error, joint task--contact macro-F1, feature/action stability under tactile noise, and action-direction cosine alignment. Each analysis uses its specified evaluation split, and learned readouts use matched head capacity across representations. Training settings are detailed in \hyperref[sec:supp_pretraining]{Appendix~\ref*{sec:supp_pretraining}} and \hyperref[sec:supp_injection]{Appendix~\ref*{sec:supp_injection}}; full analysis protocols are provided in \hyperref[sec:supp_repr]{Appendix~\ref*{sec:supp_repr}}.

\clearpage
\input{tables/closed_loop}
\subsection{Real-World Manipulation (Q1)}
\label{sec:closed_loop}
\paragraph{Manipulation performance across policies.}
WiTac improves closed-loop success over the base policy and \witacnw on all four tasks with both VLA backbones (\cref{tab:closed_loop}). With GR00T, WiTac reaches an average success rate of 66.3\%, exceeding the vision-only base policy (37.5\%) by 28.8\%, \witacnw (47.5\%) by 18.8\%, and the standalone tactile policies FTP-1 (40.0\%) and T-Rex (50.0\%) on every task. Notably, on USB reinsertion, adding wrist-free touch (\witacnw) drops success from 10\% to 5\%, whereas WiTac raises it to 40\% (+35\%), indicating that ungrounded fingertip touch can mislead fine extraction and insertion without wrist spatial context. Under the same SharpaDex v1.0 pretraining and GR00T backbone, VT-MUSE joint repr.\ lifts average success to 50.0\% over \witacnw (47.5\%), yet WiTac surpasses it by 16.3\% on average, demonstrating the effectiveness of our wrist-conditioned, finger-level contact encoder design. Reusing the frozen WiTac encoder with $\pi_{0.5}$ similarly lifts average success from 43.8\% (base) and 51.3\% (\witacnw) to 58.8\%, with consistent 5--10\% gains over \witacnw across all tasks. Qualitative rollouts for all four tasks are provided in \hyperref[sec:supp_success_failure]{Appendix~\ref*{sec:supp_success_failure}} (\cref{fig:real_robot_success_failure}).

\Needspace{13\baselineskip}
\input{tables/contact_stages}
\paragraph{Contact-related behaviors.}
We examine two critical contact stages on GR00T: battery alignment and USB extraction. \Cref{fig:contact_stage_cases} contrasts \witacnw and WiTac rollouts from matched initial conditions. \witacnw rotates the battery out of alignment and lifts its fingers without extracting the USB connector, whereas WiTac adjusts contact to seat the battery and extract the connector. These examples accompany the aggregate results in \cref{tab:contact_stages}: across 20 trials per task, WiTac reduces battery misalignment failures and improves USB extraction success.
\input{figures/contact_stage_cases}

\subsection{Representation Analysis (Q2)}
\label{sec:repr}
To understand how wrist conditioning shapes the learned tokens, we probe frozen WiTac and \witacnw across contact reconstruction, context readout, tactile-noise perturbations, and action-direction readout.

\paragraph{More accurate contact reconstruction.}
We assess three complementary aspects of contact: area describes its spatial extent, intensity measures the magnitude of tactile deformation, and center locates the contact on the sensor. We evaluate the pretrained reconstruction heads on 9,632 validation windows from 602 episodes in the SharpaDex v1.0 dataset. WiTac reduces area and intensity MAE by 7.57\% and 10.00\% over all windows, and contact-center error by 2.64\%, % (\cref{fig:contact_reconstruction})
with area and intensity errors also decreasing on contact-containing samples by 1.30\% and 5.53\%, respectively. These results support more accurate recovery of contact descriptors (full results in \cref{tab:contact_reconstruction}). The USB-reinsertion and cabinet-opening examples in \cref{fig:tactile_reconstruction} illustrate improved recovery of deformation changes.
\input{figures/tactile_reconstruction}

\input{figures/representation_pair}
\paragraph{Joint readout of task context and contact state.}
We next compare WiTac against \witacnw, wrist DINO features, and \witacnw{}+DINO concatenation on predicting both the task and its contact state. Each target is a task--state pair (e.g., battery insertion with stable contact), combining the four tasks with short-window tactile labels: the 12-class setting uses no, stable, and changing contact, while the 20-class setting adds contact onset and release. Each representation is reduced to 50 PCA dimensions and evaluated with an equal-capacity classifier using macro-F1 on the original test split and a held-out confirmation split.

WiTac improves macro-F1 over DINO by 4.39--10.14\% and over \witacnw by 1.83--7.65\% across all four evaluations (\cref{fig:joint_readout}; full scores in \cref{tab:joint_readout}). These gains support the complementary roles of visual context and tactile information in distinguishing task--contact states. Compared with \witacnw{}+DINO concatenation, the absolute differences are at most 1.03\% in three settings, with a larger 4.86\% gain on the 20-class confirmation split. Thus, while joint encoding does not consistently surpass concatenation on this readout, stronger fusion mechanisms for fine-grained contact-state discrimination remain a promising direction.

\Needspace{17\baselineskip}
\input{tables/noise_feature_action}
\paragraph{Feature stability and noisy-input action accuracy.}
\label{sec:robustness}
Real-world tactile sensors often exhibit hardware instability and measurement noise during deployment. To evaluate stability under such perturbations, we inject Gaussian noise ($\sigma=3$ in 8-bit intensity) into tactile inputs while keeping wrist images and robot state unchanged. We measure two metrics on identical perturbed windows: \emph{feature drift} $D = 1-\cos(z_{\mathrm{clean}},z_{\mathrm{noisy}})$ averaged over valid contact tokens $z$, and \emph{noisy-input action MAE} (mean joint-angle error in degrees against demonstrated actions). \Cref{tab:noise_feature_action} reports the percentage error reduction of WiTac relative to \witacnw ($G_D$ and $G_{\mathrm{MAE}}$, where positive values indicate lower drift or error than \witacnw). WiTac reduces feature drift by 35.6--44.1\% across all four tasks. Downstream action MAE also drops by 7.35--22.76\% across all four GR00T tasks, while $\pi_{0.5}$ improves on USB reinsertion and cabinet opening with minor regressions on battery insertion and bowl wiping. Overall, wrist conditioning consistently stabilizes contact features against tactile noise, whereas action-level gains depend on the downstream policy and task (full metrics in \cref{tab:noise_stability,tab:noise_action_complete}, \hyperref[sec:supp_repr]{Appendix~\ref*{sec:supp_repr}}).

\Needspace{14\baselineskip}
\input{tables/direction_readout_selected}
\paragraph{Action-direction readout.}
Similar contacts can call for different corrective motions depending on hand--object alignment. Using frozen features and equal-capacity linear probes (PCA16) fitted on separate training episodes, we predict upcoming joint-command changes $\widehat{\Delta q}_t$ and measure cosine similarity $\cos(\widehat{\Delta q}_t,\Delta q_t)$ against demonstrated short-horizon adjustments $\Delta q_t$ across 25 contact-matched test pairs per task, where paired observations share similar touch but precede distinct $\Delta q_t$ (\cref{tab:direction_readout_selected}; protocol in \hyperref[sec:supp_repr]{Appendix~\ref*{sec:supp_repr}}). Averaging cosine similarity over both observations in each pair, WiTac achieves the highest directional alignment across all four tasks. \Cref{fig:battery_context_case} illustrates a battery test pair on the 7-DoF right arm (\texttt{J1}--\texttt{J7}), where WiTac recovers both adjustment directions (0.865 and 0.736) while DINO and \witacnw fail on one.

\subsection{Ablation Studies (Q3)}
\label{sec:ablation}
Using GR00T as the policy backbone, we ablate representation pretraining, masked visual supervision, and pre-injection fusion on downstream task success. Variant definitions are provided in \hyperref[sec:supp_ablation]{Appendix~\ref*{sec:supp_ablation}}.

\Needspace{15\baselineskip}
\input{tables/pretraining_ablation}
\paragraph{Contribution of representation pretraining.}
In the variant without representation pretraining, the SharpaWave Deform Encoder remains frozen while the remaining WiTac modules are trained jointly with the policy during task fine-tuning. Pretrained WiTac achieves 5--25\% higher success across all four tasks, improving the average success rate from 53.8\% to 66.3\% (\cref{tab:pretraining_ablation}), with the largest gain on USB reinsertion (15\% to 40\%). These gains demonstrate the effectiveness of our representation pretraining prior to policy learning.

\paragraph{Effect of masked visual recovery.}
To test whether cross-attending to wrist features suffices without visual supervision, we pretrain \emph{WiTac w/o visual recovery} on SharpaDex v1.0 by keeping wrist inputs and all contact, motion, and wrench losses while removing masked visual recovery ($D_{\mathrm{vis}}$). Although wrist views without $D_{\mathrm{vis}}$ raise average success to 58.8\% over \witacnw (47.5\%) and unpretrained WiTac (53.8\%), omitting $D_{\mathrm{vis}}$ reduces average success by 7.5\% versus full WiTac (66.3\%, \cref{tab:pretraining_ablation}), with the sharpest drops on fine-alignment tasks such as USB reinsertion (25\% vs.\ 40\%) and battery insertion (45\% vs.\ 55\%). Supervising masked wrist recovery thus keeps the ten-token bottleneck grounded in visual context.

\paragraph{Joint encoding also exceeds separate injection.}
Both joint encoding and separate injection supply tactile and wrist visual features to the policy, differing in whether the two modalities enter the policy through independent attention branches or are fused beforehand into unified contact tokens. Fusing them before policy injection yields 10--25\% higher success across all four tasks, raising the four-task average from 50.0\% to 66.3\% (\cref{tab:pretraining_ablation}). This confirms that grounding touch in wrist visual context prior to policy injection is substantially more effective than leaving cross-modal fusion to the action policy.
% The separate-injection and joint-encoding configurations differ in their token budgets and interface structure, so this comparison measures the benefit of the complete fusion design rather than an isolated change under matched compute.

%% file: tables/closed_loop.tex
\begin{table}[t]
\centering
\caption{\textbf{Closed-loop success rates (\%).} Average is the unweighted mean across all four tasks. \witacnw denotes WiTac without wrist conditioning.}
\label{tab:closed_loop}
\small
\setlength{\tabcolsep}{6pt}
\begin{adjustbox}{max width=\linewidth}
\begin{tabular*}{\textwidth}{@{\extracolsep{\fill}}lccccc@{}}
\toprule
Policy & Insert battery & Wipe bowl & Reinsert USB & Open cabinet & Average \\
\midrule
FTP-1~\cite{ftp1_2026} & 40 & 55 & 5 & 60 & 40.0 \\
T-Rex~\cite{trex2026} & 45 & 65 & 15 & 75 & 50.0 \\
\midrule
GR00T N1.7~\cite{gr00tn1_2025} & 30 & 40 & 10 & 70 & 37.5 \\
+ \witacnw & 45 & 60 & 5 & 80 & 47.5 \\
% + Separate injection & 40 & 65 & 15 & 80 & 50.0 \\
+ VT-MUSE~\cite{vtmuse2026} joint repr. & 40 & 70 & 15 & 75 & 50.0 \\
\textbf{+ WiTac} & \textbf{55} & \textbf{80} & \textbf{40} & \textbf{90} & \textbf{66.3} \\
\midrule
$\pi_{0.5}$~\cite{pi05_2025} & 55 & 65 & 10 & 45 & 43.8 \\
+ \witacnw & 60 & 65 & 15 & 65 & 51.3 \\
\textbf{+ WiTac} & \textbf{70} & \textbf{75} & \textbf{20} & \textbf{70} & \textbf{58.8} \\
\bottomrule
\end{tabular*}
\end{adjustbox}
\end{table}

%% file: tables/contact_stages.tex
\begin{wraptable}{r}{0.48\textwidth}
\centering
\captionsetup{font=footnotesize,skip=4pt}
\caption{\textbf{Contact-stage outcomes with GR00T (20 trials per task).} Misalignment failures during battery insertion and partial-task extraction success on USB reinsertion.}
\label{tab:contact_stages}
\footnotesize
\setlength{\tabcolsep}{4pt}
\begin{adjustbox}{max width=\linewidth}
\begin{tabular}{lcc}
\toprule
Method & \shortstack{Misalignment\\failures $\downarrow$} & \shortstack{Extraction\\success (\%) $\uparrow$} \\
\midrule
GR00T & 7 & 50 \\
GR00T + \witacnw & 4 & 60 \\
% GR00T + Separate injection & 4 & 75 \\
\textbf{GR00T + WiTac} & \textbf{1} & \textbf{95} \\
\bottomrule
\end{tabular}
\end{adjustbox}
\end{wraptable}

%% file: figures/contact_stage_cases.tex
\begin{figure}[t]
\centering
\includegraphics[width=\textwidth]{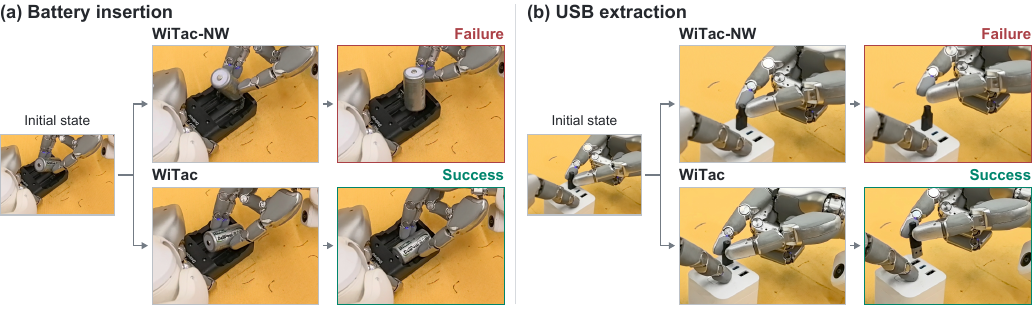}
\caption{\textbf{Contact-stage rollouts from matched initial conditions.} Sequences progress left to right. In these examples, \witacnw rotates the battery out of alignment and leaves the USB connector inserted, whereas WiTac seats the battery and extracts the connector.}
\label{fig:contact_stage_cases}
\end{figure}

%% file: figures/tactile_reconstruction.tex
\begin{figure}[t]
    \centering
    \includegraphics[width=0.90\textwidth]{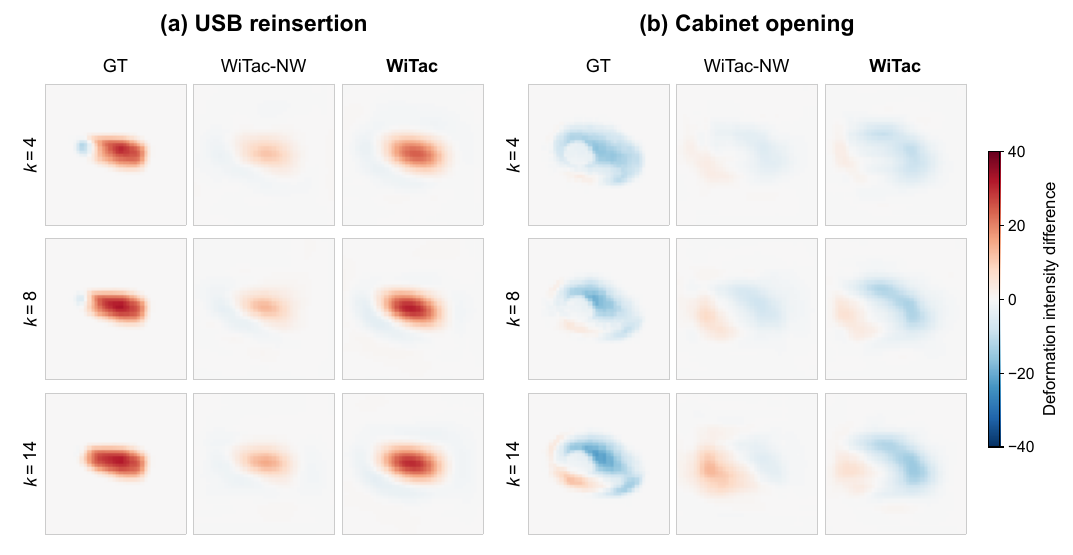}
    \caption{\textbf{Qualitative examples of contact-change reconstruction.} Right-index deformation differences $T_t-T_{t-k}$ at offsets $k\in\{4,8,14\}$ frames. Methods share a fixed crop within each case and a signed color scale.}
    \label{fig:tactile_reconstruction}
\end{figure}

%% file: figures/representation_pair.tex
\begin{figure}[t]
\centering
\begin{minipage}[t]{0.48\linewidth}
\centering
\begin{minipage}[c][5.2cm][c]{\linewidth}
\centering
\includegraphics[width=\linewidth,height=5.2cm,keepaspectratio]{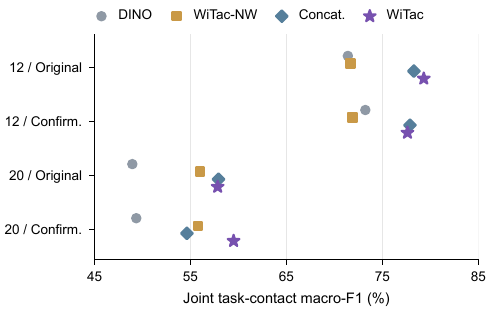}
\end{minipage}
\captionsetup{font=footnotesize}
\caption{\textbf{Joint task--contact readout.} Macro-F1 (\%) over four tasks with three or five contact states, giving 12 or 20 classes. Original/Confirm. identify the two test splits; Concat. denotes \witacnw{}+DINO.}
\label{fig:joint_readout}
\end{minipage}
\hfill
\begin{minipage}[t]{0.48\linewidth}
\centering
\begin{minipage}[c][5.2cm][c]{\linewidth}
\centering
\includegraphics[width=\linewidth,height=5.2cm,keepaspectratio]{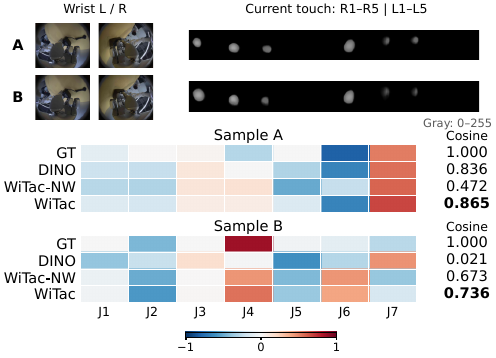}
\end{minipage}
\captionsetup{font=footnotesize}
\caption{\textbf{Action-direction readout on a battery test pair.} Top: wrist views and current touch for observations A and B with similar contact patterns. Bottom: unit-normalized GT and linear-probe right-arm joint adjustments ($\Delta q_t/\lVert\Delta q_t\rVert_2$, \texttt{J1}--\texttt{J7}), with cosine similarity to GT on the right.}
    \label{fig:battery_context_case}
\end{minipage}
\end{figure}

%% file: tables/noise_feature_action.tex
\begin{wraptable}{r}{0.46\textwidth}
\centering
\captionsetup{font=footnotesize,skip=4pt}
\caption{\textbf{Feature stability improves; action gains vary.} $G_D$ (shared across policies for the frozen encoder) and $G_{\mathrm{MAE}}$ are relative reductions (\%) in feature drift and noisy-input action MAE versus \witacnw under Gaussian noise ($\sigma=3$). Positive values indicate improvement.}
\label{tab:noise_feature_action}
\footnotesize
\setlength{\tabcolsep}{5.5pt}
\begin{adjustbox}{max width=\linewidth}
\begin{tabular}{lrrr}
\toprule
Task & $G_D$ & \multicolumn{2}{c}{$G_{\mathrm{MAE}}$} \\
\cmidrule(lr){3-4}
& & GR00T & $\pi_{0.5}$ \\
\midrule
Insert battery & $+37.4$ & $+22.76$ & $-0.56$ \\
Wipe bowl & $+35.6$ & $+7.35$ & $-1.90$ \\
Reinsert USB & $+37.0$ & $+9.18$ & $+1.91$ \\
Open cabinet & $+44.1$ & $+21.59$ & $+2.19$ \\
\bottomrule
\end{tabular}
\end{adjustbox}
\end{wraptable}

%% file: tables/direction_readout_selected.tex
\begin{wraptable}{r}{0.46\textwidth}
\centering
\captionsetup{font=footnotesize,skip=4pt}
\caption{\textbf{Action-direction readout on matched test pairs.} Mean cosine similarity $\cos(\widehat{\Delta q}_t,\Delta q_t)$ $\uparrow$ across 25 test pairs per task.}
\label{tab:direction_readout_selected}
\footnotesize
\setlength{\tabcolsep}{4pt}
\begin{adjustbox}{max width=\linewidth}
\begin{tabular}{lrrr}
\toprule
Task & DINO & \witacnw & WiTac \\
\midrule
Insert battery & 0.6814 & 0.6794 & \textbf{0.7325} \\
Reinsert USB & 0.3281 & 0.3621 & \textbf{0.4295} \\
Wipe bowl & 0.7246 & 0.7696 & \textbf{0.7881} \\
Open cabinet & 0.6993 & 0.6669 & \textbf{0.7386} \\
\bottomrule
\end{tabular}
\end{adjustbox}
\end{wraptable}

%% file: tables/pretraining_ablation.tex
\begin{wraptable}{r}{0.54\textwidth}
\centering
\captionsetup{font=footnotesize,skip=4pt}
\caption{\textbf{Design ablations on GR00T.} Closed-loop success rates (\%); full WiTac combines representation pretraining, masked visual recovery, and pre-injection joint encoding.}
\label{tab:pretraining_ablation}
\footnotesize
\setlength{\tabcolsep}{2.5pt}
\begin{adjustbox}{max width=\linewidth}
\begin{tabular}{lccccc}
\toprule
Variant & \shortstack{Insert\\battery} & \shortstack{Wipe\\bowl} & \shortstack{Reinsert\\USB} & \shortstack{Open\\cabinet} & Average \\
\midrule
WiTac w/o pretraining & 45 & 75 & 15 & 80 & 53.8 \\
WiTac w/o visual recovery & 45 & \textbf{80} & 25 & 85 & 58.8 \\
Separate injection & 40 & 65 & 15 & 80 & 50.0 \\
\textbf{Joint (WiTac)} & \textbf{55} & \textbf{80} & \textbf{40} & \textbf{90} & \textbf{66.3} \\
\bottomrule
\end{tabular}
\end{adjustbox}
\end{wraptable}

%% file: sections/5_conclusion.tex
\section{Conclusion}
\label{sec:conclusion}
We presented WiTac, a pretrained encoder that integrates tactile histories, hand kinematics, and wrist vision into finger-level contact tokens before action generation. Across four contact-rich tasks, WiTac improves average success over its wrist-free variant \witacnw by 18.8\% and 7.5\% with GR00T N1.7 and \(\pi_{0.5}\), respectively. Offline analyses further show that WiTac reconstructs contact descriptors more accurately, better distinguishes joint task--contact states, exhibits lower feature drift under synthetic tactile noise, and better aligns with demonstrated adjustment directions than \witacnw. Together, these findings support incorporating wrist context into tactile representations for dexterous manipulation.

\paragraph{Limitations.}
WiTac uses feature-level cross-attention without depth-based alignment of vision and touch in a shared 3D space, despite incorporating fingertip kinematics. It combines tactile histories with only current wrist images, without explicitly modeling their joint temporal evolution. Our evaluation is also limited to four tasks on one robotic platform; transfer across hand morphologies, tactile sensors, and camera configurations remains to be validated.

%% file: appendix.tex
% Compact appendix float and heading spacing.
\captionsetup[table]{font=footnotesize,skip=3.5pt}
\captionsetup[figure]{font=footnotesize,skip=4.5pt}
\setlength{\floatsep}{7pt plus 2pt minus 2pt}
\setlength{\textfloatsep}{9pt plus 2pt minus 2pt}
\setlength{\intextsep}{7pt plus 2pt minus 2pt}
\makeatletter
\setlength{\@fptop}{0pt}
\setlength{\@fpsep}{9pt}
\setlength{\@fpbot}{0pt plus 1fil}
\makeatother
\titlespacing*{\section}{0pt}{8pt plus 2pt minus 1pt}{2pt}
\titlespacing*{\subsection}{0pt}{6pt plus 2pt minus 1pt}{1.5pt}
\titlespacing*{\paragraph}{0pt}{3pt plus 1pt minus 1pt}{0.75em}
\setlist[itemize]{itemsep=1pt,topsep=1.5pt,parsep=0pt}
% Shared appendix content, used by the combined paper and optional standalone copy.
\appendix
\setcounter{table}{0}
\setcounter{figure}{0}
\renewcommand{\thetable}{S\arabic{table}}
\renewcommand{\thefigure}{S\arabic{figure}}
% Unique hyperlink destinations after resetting figure/table counters.
\renewcommand{\theHtable}{supp.\arabic{table}}
\renewcommand{\theHfigure}{supp.\arabic{figure}}
\input{sections/supp_method}

\input{sections/supp_experiments}
\FloatBarrier

%% file: sections/supp_method.tex
\section{Method and Implementation Details}
\label{sec:supp_method}

This section supplements Section~3 of the main paper with the temporal offsets, objective weights, and training hyperparameters for WiTac and downstream policy fine-tuning.

\subsection{Contact Encoding and Pretraining (Main Sec.~3.1 \& 3.2)}
\label{sec:supp_pretraining}

At each decision step $t$ (30\,Hz), the encoder takes an eight-frame tactile history $\mathcal{T}_t$ at frame offsets $(-14, -12, -10, -8, -6, -4, -2, 0)$ across all ten fingers, together with the 65D robot joint state $s_t$, wrist-relative fingertip poses $p_t$ from forward kinematics, and dual-wrist fisheye RGB images $I_t$ resized to $256\times320$ and encoded by a frozen DINOv3 ViT-B/16~\cite{dinov3_2025}. The spatial tactile encoder is initialized from the SharpaWave Deform Encoder~\cite{sharpawave_deform_encoder2026}, followed by causal temporal convolutions, FiLM pose modulation, and two Transformer fusion blocks (8 heads, width 768) that output ten contact tokens $C_t\in\mathbb{R}^{10\times768}$. The wrist-free variant \witacnw uses the same tactile, state, and kinematic encoding architecture, omitting only the wrist cross-attention branch and masked visual recovery objective.

\Cref{tab:pretraining_targets} lists the temporal offsets, prediction horizons, and loss weights $\lambda_j$ for the five self-supervised pretraining objectives in Eq.~(5) of the main paper. In deformation-change reconstruction ($D_{\mathrm{def}}$), changing and static pixels are weighted equally in a balanced L1 loss so that static background regions do not dominate training. In masked visual recovery ($D_{\mathrm{vis}}$), 50\% of patch blocks within the hand--object wrist regions are masked before ViT encoding to form $V_t^{\mathrm{vis}}$, and the decoder reconstructs the LayerNorm-normalized full-image DINOv3 features at masked positions. Pretraining hyperparameters on SharpaDex v1.0~\cite{sharpadex_v1_2026} are summarized in \cref{tab:pretraining_hparams}.

\begin{table}[!htbp]
\centering
\caption{\textbf{Pretraining objectives, temporal offsets, and loss weights (Main Sec.~3.2).} Frame offsets and horizons are defined at 30\,Hz. All prediction heads decode the shared contact tokens $C_t$ during pretraining and are discarded at deployment.}
\label{tab:pretraining_targets}
\footnotesize
\setlength{\tabcolsep}{3.5pt}
\begin{tabular*}{\linewidth}{@{\extracolsep{\fill}}llll@{}}
\toprule
Objective & Target & Offsets / horizons & Loss \& weight ($\lambda_j$) \\
\midrule
Contact descriptor ($D_{\mathrm{desc}}$) & Contact area, intensity, 2D center & Current frame ($t$) & Smooth L1 ($0.05$) \\
Deformation change ($D_{\mathrm{def}}$) & $\Delta T_{t,k}=T_t-T_{t-k}$ ($64\times64$) & $k\in\{4, 8, 14\}$ past frames & Balanced L1 ($0.10$) \\
Fingertip motion ($D_{\mathrm{mot}}$) & Wrist-frame $\Delta x_{t,k}=x_t-x_{t-k}$ & $k\in\{2, 6, 10, 14\}$ past frames & Smooth L1 ($0.05$) \\
Visual recovery ($D_{\mathrm{vis}}$) & Masked full-image DINOv3 features & Current frame ($t$) & LayerNorm MSE ($0.10$) \\
Wrench response ($D_{\mathrm{wr}}$) & 6D wrench change $\Delta w_{t,h}=w_{t+h}-w_t$ & $h\in\{4, 8, 16\}$ future frames & Smooth L1 ($0.10$) \\
\bottomrule
\end{tabular*}
\end{table}

\subsection{Policy Injection and Fine-Tuning (Main Sec.~3.3)}
\label{sec:supp_injection}

During downstream policy learning, the pretrained contact encoder (WiTac or \witacnw) is frozen and injected into alternating action-expert blocks of GR00T N1.7~\cite{gr00tn1_2025} and $\pi_{0.5}$~\cite{pi05_2025} via gated cross-attention ($\lambda_{\mathrm{gate}}=0.02$). Both policies receive head stereo RGB ($512\times640$), dual-wrist fisheye RGB ($384\times480$ for the native policy visual stream; the same dual-wrist fisheye feeds are resized to $256\times320$ in native orientation for WiTac), language instructions, and 65D robot joint states, and predict relative joint-angle action chunks. Binary contact-gate labels $y_t\in\{0,1\}$ are derived from the active-pixel fraction on the primary manipulation fingers (thumb, index, and middle of both hands). \Cref{tab:policy_hparams} summarizes the downstream fine-tuning hyperparameters; on $\pi_{0.5}$, the pretrained policy backbone is fine-tuned with learning rate $2.5\times10^{-5}$ while the newly introduced contact cross-attention and gate modules use $10^{-4}$.

\input{tables/training_settings_pair}

%% file: tables/training_settings_pair.tex
\begin{table}[!htbp]
\centering
\begin{minipage}[t]{0.485\linewidth}
\captionsetup{font=footnotesize,skip=4pt}
\input{tables/pretraining_hparams}
\end{minipage}
\hfill
\begin{minipage}[t]{0.485\linewidth}
\captionsetup{font=footnotesize,skip=4pt}
\input{tables/policy_hparams}
\end{minipage}
\end{table}

%% file: tables/pretraining_hparams.tex
\centering
\caption{\textbf{Representation pretraining hyperparameters (Main Sec.~3.2).}}
\label{tab:pretraining_hparams}
\footnotesize
\setlength{\tabcolsep}{2pt}
\begin{adjustbox}{max width=\linewidth}
\begin{tabular}{@{}>{\raggedright\arraybackslash}p{0.29\linewidth}>{\raggedright\arraybackslash}p{\dimexpr0.71\linewidth-4pt\relax}@{}}
\toprule
Setting & Value \\
\midrule
Dataset & SharpaDex v1.0~\cite{sharpadex_v1_2026} (59 tasks) \\
Train / val split & 98\% / 2\% episodes \\
Training steps & 30,000 \\
Global batch size & 1,024 (\texttt{bfloat16}) \\
Optimizer & AdamW (weight decay $10^{-3}$, grad clip $1.0$) \\
Learning rate & $10^{-5}$ (spatial backbone), $10^{-4}$ (other modules) \\
LR schedule & 2,000-step warmup, cosine decay to $0.1\times$ \\
\bottomrule
\end{tabular}
\end{adjustbox}

%% file: tables/policy_hparams.tex
\centering
\caption{\textbf{Policy fine-tuning settings (Main Sec.~3.3 \& 4.1).} $\pi_{0.5}$ LR: policy backbone / contact interface.}
\label{tab:policy_hparams}
\footnotesize
\setlength{\tabcolsep}{2pt}
\begin{adjustbox}{max width=\linewidth}
\begin{tabular}{@{}>{\raggedright\arraybackslash}p{0.30\linewidth}>{\raggedright\arraybackslash}p{0.34\linewidth}>{\raggedright\arraybackslash}p{\dimexpr0.36\linewidth-8pt\relax}@{}}
\toprule
Setting & GR00T N1.7~\cite{gr00tn1_2025} & $\pi_{0.5}$~\cite{pi05_2025} \\
\midrule
Action horizon & 40 steps & 25 steps \\
Demonstrations & 200 per task & 200 per task \\
Training schedule & 30 epochs & 30 epochs \\
Optimizer & AdamW & AdamW \\
Learning rate & $10^{-4}$ & $2.5\times10^{-5}$ / $10^{-4}$ \\
Warmup \& decay & 2,000 steps, cosine & 1,000 steps, cosine \\
\bottomrule
\end{tabular}
\end{adjustbox}

%% file: sections/supp_experiments.tex
\section{Additional Experimental Details and Results}
\label{sec:supp_exp}

This section supplements Section~4 of the main paper with task and baseline descriptions (Sec.~4.1), qualitative real-world rollouts (Sec.~4.2), complete numerical tables for the representation analyses (Sec.~4.3), and ablation setups (Sec.~4.4).

\subsection{Tasks and Baselines (Main Sec.~4.1)}
\label{sec:supp_setup}

\paragraph{Task definitions.}
We evaluate four bimanual contact-rich manipulation tasks (none of which are included in the 59-task SharpaDex v1.0 pretraining dataset) over 20 real-world trials per method and task, with all policies fine-tuned for 30 epochs on 200 teleoperated demonstrations per task (recorded at 30\,Hz):
\begin{itemize}
\item \textbf{Insert battery:} One hand stabilizes the battery holder while the other grasps a cylindrical battery, aligns it with the compartment slot under contact, and presses it down until fully seated.
\item \textbf{Wipe bowl:} One hand stabilizes a bowl while the other grasps a sponge and maintains continuous interior contact to wipe the inner wall.
\item \textbf{Reinsert USB:} One hand stabilizes a power strip while the other grasps and extracts a plugged connector, re-aligns the plug with a target socket, and presses it in until seated.
\item \textbf{Open cabinet:} The robot grasps and turns a key in the cabinet lock to unlatch it, then pulls the articulated door open along its hinge arc.
\end{itemize}

\paragraph{External baseline adaptations.}
All external baselines are fine-tuned on the same 200 demonstrations per task for 30 epochs:
\begin{itemize}
\item \textbf{VT-MUSE joint repr.~\cite{vtmuse2026}:} We adapt the sequential visuotactile encoder of VT-MUSE to eight-frame histories of the ten tactile sensors and two wrist cameras, pretrain it on the same SharpaDex v1.0 split using its two-stage alignment and masked reconstruction objectives, and inject the frozen encoder into GR00T N1.7 via the same gated cross-attention interface as WiTac.
\item \textbf{FTP-1~\cite{ftp1_2026} and T-Rex~\cite{trex2026}:} Fine-tuned with global batch size 256 from the pretrained checkpoint for FTP-1 and the mid-trained checkpoint for T-Rex, adapting their respective visual, tactile, and proprioceptive input pipelines to the Sharpa North observation and relative joint-action space.
\end{itemize}

\subsection{Qualitative Real-World Rollouts (Main Sec.~4.2)}
\label{sec:supp_success_failure}
\Cref{fig:real_robot_success_failure} compares successful WiTac rollouts with failure rollouts from \witacnw on GR00T N1.7 across all four real-world tasks, complementing the closed-loop success rates and contact-stage behaviors reported in Section~4.2 of the main paper.
\input{figures/real_robot_success_failure}

\subsection{Complete Representation Analysis Results (Main Sec.~4.3)}
\label{sec:supp_repr}

\paragraph{Contact reconstruction.}
\Cref{tab:contact_reconstruction} reports the complete absolute errors and relative reductions on the 9,632 SharpaDex v1.0 validation windows corresponding to the contact-reconstruction analysis in Section~4.3, alongside the qualitative deformation-change reconstructions illustrated in Figure~4 of the main paper.

\paragraph{Joint task--contact readout.}
\Cref{tab:joint_readout} provides the complete macro-F1 scores and sample counts corresponding to Figure~5 of the main paper. Short-window contact states are categorized from the eight-frame tactile history into \emph{no contact}, \emph{onset}, \emph{release}, \emph{stable contact}, and \emph{changing contact}. The 12-class setting pairs the four tasks with \{no, stable, changing\} contact, while the 20-class setting uses all five contact states. Frozen representations are reduced to 50 PCA dimensions and evaluated with a class-balanced multinomial logistic regression probe.

\input{tables/representation_results_pair}

\paragraph{Synthetic tactile-noise perturbations.}
\Cref{tab:noise_stability,tab:noise_action_complete} provide the full numerical metrics supplementing Table~3 of the main paper under Gaussian tactile noise ($\sigma=3$ in 0--255 intensity units, with wrist images and robot state held fixed). Alongside feature drift $1-\cos(z_{\mathrm{clean}},z_{\mathrm{noisy}})$ and action MAE (averaged over the first 25 command steps), \cref{tab:noise_stability} reports cross-attention Jensen--Shannon (JS) divergence and the relative $\ell_2$ drift of the pre-gate contact-attention output $r$, defined as $\lVert r_{\mathrm{noisy}}-r_{\mathrm{clean}}\rVert_2 / (\lVert r_{\mathrm{clean}}\rVert_2+10^{-8})$. WiTac reduces feature drift by 35.6--44.1\% across all four tasks, as well as attention JS by 26.8--79.7\% and contact-output drift by 15.7--28.8\% across all eight policy--task pairs.

\input{tables/noise_stability}
\input{tables/noise_action_complete}

\paragraph{Action-direction readout.}
In Table~4 and Figure~6 of the main paper, demonstrated adjustment targets $\Delta q_t$ are defined over a 16-step future command window (the difference between the mean of the last four and first four joint commands). Frozen DINO, \witacnw, and WiTac features are standardized on training episodes, projected to 16 whitened PCA components, and mapped to $\widehat{\Delta q}_t$ via ridge-regression linear probes with regularization selected by three-fold cross-validation over training episodes. Directional alignment is evaluated by $\cos(\widehat{\Delta q}_t,\Delta q_t) = \widehat{\Delta q}_t^\top \Delta q_t / (\lVert\widehat{\Delta q}_t\rVert_2 \lVert\Delta q_t\rVert_2)$, averaged over the two contact-matched observations in each held-out test pair and across the 25 test pairs per task.

\subsection{Ablation Setup (Main Sec.~4.4)}
\label{sec:supp_ablation}

Section~4.4 (Table~5 of the main paper) compares full pretrained WiTac against three design variants on GR00T N1.7 under the same 30-epoch training schedule and 200 demonstrations per task:
\begin{itemize}
\item \textbf{WiTac w/o pretraining:} The SharpaWave Deform Encoder remains frozen while the remaining WiTac modules are trained end-to-end with the GR00T policy and gated contact interface during task fine-tuning, without self-supervised pretraining on SharpaDex v1.0.
\item \textbf{WiTac w/o visual recovery:} Pretrained on the same SharpaDex v1.0 split with dual-wrist cross-attention and all tactile, motion, and wrench objectives ($D_{\mathrm{desc}}, D_{\mathrm{def}}, D_{\mathrm{mot}}, D_{\mathrm{wr}}$), omitting only the masked visual recovery loss ($D_{\mathrm{vis}}$) before freezing the encoder for GR00T fine-tuning.
\item \textbf{Separate injection:} The frozen \witacnw encoder and frozen DINOv3 wrist patch tokens are injected into the GR00T action expert through two parallel cross-attention branches instead of fusing wrist vision into contact tokens prior to policy injection.
\end{itemize}

%% file: figures/real_robot_success_failure.tex
\begin{figure}[p]
\centering
\definecolor{rolloutgreen}{RGB}{55,133,105}
\definecolor{rolloutred}{RGB}{186,98,98}
% Strip height is derived from \textheight so the four tasks fill one float page.
\newlength{\rolloutstrip}
\setlength{\rolloutstrip}{\dimexpr(\textheight-128pt)/8\relax}
\newcommand{\rolloutlabel}[2]{%
  \parbox[c]{1.6cm}{\raggedright\scriptsize\bfseries\textcolor{#1}{#2}}}
\newcommand{\rolloutrow}[3]{%
  \rolloutlabel{#1}{#2}\hspace{4pt}%
  \raisebox{-0.5\height}{\includegraphics[height=\rolloutstrip,viewport=#3,clip]{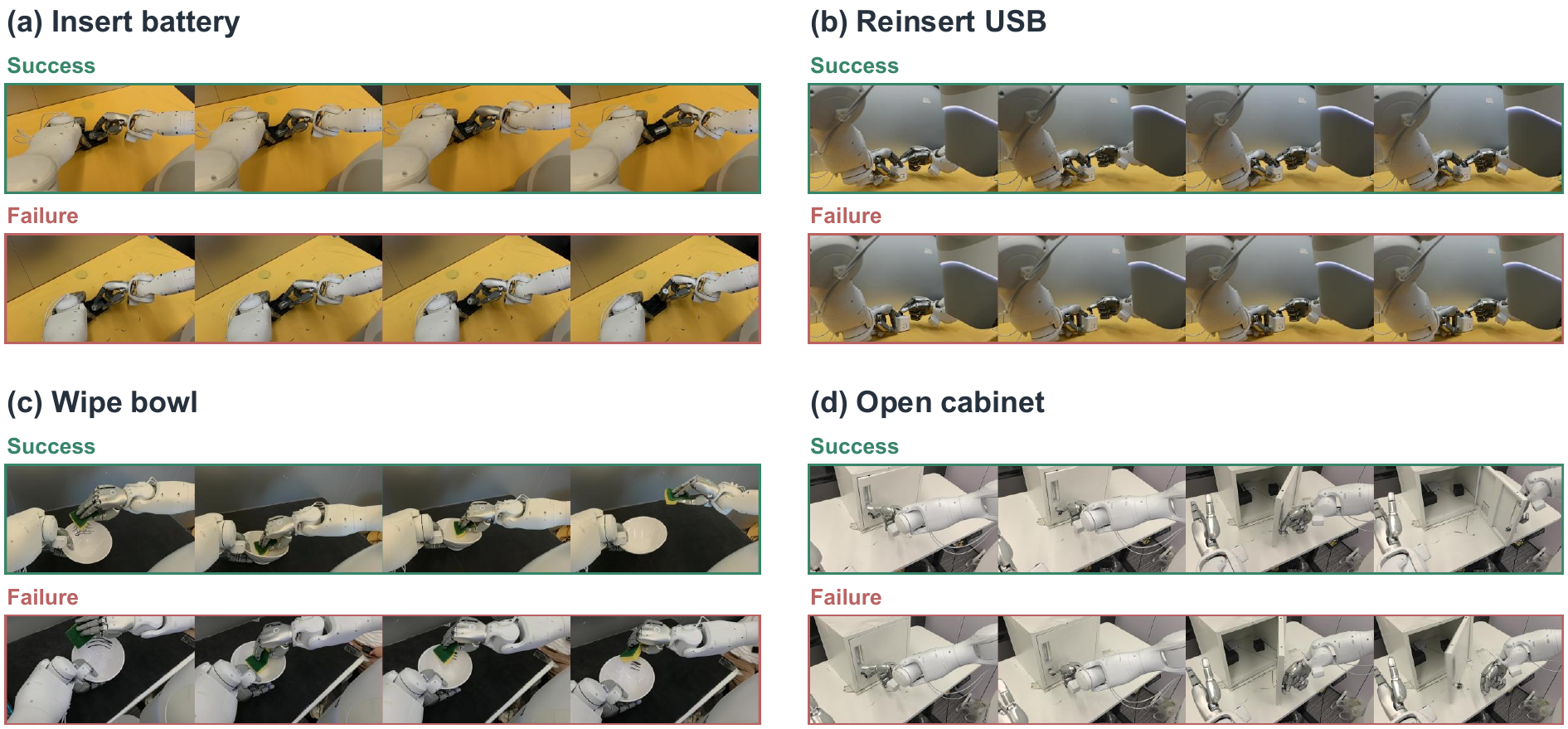}}}
\newcommand{\rollouttask}[3]{%
  \begin{minipage}{\linewidth}
    \centering
    \makebox[\linewidth][l]{\footnotesize\bfseries #1}\par\vspace{2pt}
    \rolloutrow{rolloutgreen}{WiTac\\(Success)}{#2}\par\vspace{2pt}
    \rolloutrow{rolloutred}{\witacnw\\(Failure)}{#3}\par
  \end{minipage}%
}
\rollouttask{(a) Insert battery}{0 307 442 376}{0 220 442 289}\par\vspace{7pt}
\rollouttask{(b) Reinsert USB}{465.5 307 907.5 376}{465.5 220 907.5 289}\par\vspace{7pt}
\rollouttask{(c) Wipe bowl}{0 87 442 156}{0 0 442 69}\par\vspace{7pt}
\rollouttask{(d) Open cabinet}{465.5 87 907.5 156}{465.5 0 907.5 69}
\caption{\textbf{Qualitative real-world rollouts comparing WiTac and \witacnw across all four tasks.}
(a) Battery insertion, (b) USB reinsertion, (c) bowl wiping, and (d) cabinet opening. For each task, the top row shows a successful rollout with WiTac (green) and the bottom row shows a failure rollout with \witacnw (red) on GR00T N1.7, with four frames ordered temporally from left to right.}
\label{fig:real_robot_success_failure}
\end{figure}

%% file: tables/representation_results_pair.tex
\begin{table}[!htbp]
\centering
\begin{minipage}[t]{0.485\linewidth}
\captionsetup{font=footnotesize,skip=4pt}
\input{tables/contact_reconstruction_content}
\end{minipage}
\hfill
\begin{minipage}[t]{0.485\linewidth}
\captionsetup{font=footnotesize,skip=4pt}
\input{tables/joint_readout_content}
\end{minipage}
\end{table}

%% file: tables/contact_reconstruction_content.tex
\centering
\caption{\textbf{WiTac improves contact-descriptor reconstruction.} Rel. denotes relative error reduction.}
\label{tab:contact_reconstruction}
\footnotesize
\setlength{\tabcolsep}{3pt}
\begin{tabular*}{\linewidth}{@{\extracolsep{\fill}}lrrr@{}}
\toprule
Error $\downarrow$ (units) & \witacnw & WiTac & Rel. (\%) \\
\midrule
Area, all (\%) & 0.3948 & \textbf{0.3649} & 7.57 \\
Area, contact (\%) & 0.7723 & \textbf{0.7623} & 1.30 \\
Intensity, all (0--255) & 0.3706 & \textbf{0.3335} & 10.00 \\
Intensity, contact (0--255) & 0.7406 & \textbf{0.6997} & 5.53 \\
Center (pixels) & 6.5204 & \textbf{6.3481} & 2.64 \\
\bottomrule
\end{tabular*}

%% file: tables/joint_readout_content.tex
\centering
\caption{\textbf{Joint task--contact macro-F1 $\uparrow$.} O/C: original/confirmation split; Concat.: \witacnw{}+DINO; $N$: windows.}
\label{tab:joint_readout}
\footnotesize
\setlength{\tabcolsep}{2.5pt}
\begin{tabular*}{\linewidth}{@{\extracolsep{\fill}}lrrrrr@{}}
\toprule
Classes/split & $N$ & DINO & \witacnw & Concat. & WiTac \\
\midrule
12 / O & 3,487 & .7140 & .7166 & .7828 & \textbf{.7931} \\
12 / C & 3,233 & .7322 & .7189 & \textbf{.7787} & .7761 \\
20 / O & 3,555 & .4894 & .5600 & \textbf{.5792} & .5783 \\
20 / C & 3,288 & .4935 & .5578 & .5463 & \textbf{.5949} \\
\bottomrule
\end{tabular*}

%% file: tables/noise_stability.tex
\begin{table}[!htbp]
\centering
\caption{\textbf{Synthetic-noise stability at Gaussian $\sigma=3$.} Feature drift depends only on the frozen encoder and task; JS denotes cross-attention Jensen--Shannon divergence, and contact-output drift is relative $\ell_2$ change before gating.}
\label{tab:noise_stability}
\footnotesize
\setlength{\tabcolsep}{4pt}
\begin{tabular*}{\linewidth}{l@{\extracolsep{\fill}}r@{\extracolsep{10pt}}r@{\extracolsep{\fill}}r@{\extracolsep{10pt}}r@{\extracolsep{\fill}}r@{\extracolsep{10pt}}r@{}}
\toprule
Policy / task & \multicolumn{2}{c}{Feature drift $\downarrow$} & \multicolumn{2}{c}{Attention JS $\downarrow$} & \multicolumn{2}{c}{Contact-output drift $\downarrow$} \\
\cmidrule(lr){2-3}\cmidrule(lr){4-5}\cmidrule(lr){6-7}
& \witacnw & WiTac & \witacnw & WiTac & \witacnw & WiTac \\
\midrule
GR00T / Insert battery & 0.5078 & \textbf{0.3177} & 0.1626 & \textbf{0.0658} & 1.4321 & \textbf{1.1506} \\
GR00T / Wipe bowl & 0.5413 & \textbf{0.3484} & 0.1437 & \textbf{0.0878} & 1.1974 & \textbf{0.8520} \\
GR00T / Reinsert USB & 0.4966 & \textbf{0.3130} & 0.1711 & \textbf{0.0348} & 1.6406 & \textbf{1.2741} \\
GR00T / Open cabinet & 0.5415 & \textbf{0.3028} & 0.2133 & \textbf{0.1560} & 1.4558 & \textbf{1.1163} \\
$\pi_{0.5}$ / Insert battery & 0.5078 & \textbf{0.3177} & 0.1023 & \textbf{0.0353} & 1.0884 & \textbf{0.8215} \\
$\pi_{0.5}$ / Wipe bowl & 0.5413 & \textbf{0.3484} & 0.0584 & \textbf{0.0289} & 1.3652 & \textbf{1.1510} \\
$\pi_{0.5}$ / Reinsert USB & 0.4966 & \textbf{0.3130} & 0.1059 & \textbf{0.0384} & 1.1275 & \textbf{0.8561} \\
$\pi_{0.5}$ / Open cabinet & 0.5415 & \textbf{0.3028} & 0.0789 & \textbf{0.0248} & 1.6630 & \textbf{1.3620} \\
\bottomrule
\end{tabular*}
\end{table}

%% file: tables/noise_action_complete.tex
\begin{table}[!htbp]
\centering
\caption{\textbf{Action error and noise-induced degradation at Gaussian $\sigma=3$.} $\Delta$MAE is noisy minus clean MAE; positive reduction favors WiTac.}
\label{tab:noise_action_complete}
\footnotesize
\setlength{\tabcolsep}{3.5pt}
\begin{tabular*}{\linewidth}{l@{\extracolsep{\fill}}r@{\extracolsep{8pt}}r@{\extracolsep{\fill}}r@{\extracolsep{8pt}}r@{\extracolsep{\fill}}r@{\extracolsep{8pt}}r@{\extracolsep{\fill}}l@{}}
\toprule
Policy / task & \multicolumn{2}{c}{Clean MAE ($^\circ$)} & \multicolumn{2}{c}{Noisy MAE ($^\circ$)} & \multicolumn{2}{c}{$\Delta$MAE ($^\circ$)} & \shortstack[l]{Reduction in $\Delta$MAE\\95\% interval ($^\circ$)} \\
\cmidrule(lr){2-3}\cmidrule(lr){4-5}\cmidrule(lr){6-7}
& \witacnw & WiTac & \witacnw & WiTac & \witacnw & WiTac & \\
\midrule
GR00T / Insert battery & \textbf{0.8785} & 0.8931 & 1.4027 & \textbf{1.0835} & 0.5242 & \textbf{0.1904} & [+0.2877, +0.3805] \\
GR00T / Wipe bowl & \textbf{0.7757} & 0.7801 & 0.9709 & \textbf{0.8996} & 0.1952 & \textbf{0.1195} & [+0.0393, +0.1198] \\
GR00T / Reinsert USB & 0.7787 & \textbf{0.7592} & 0.9950 & \textbf{0.9037} & 0.2163 & \textbf{0.1445} & [+0.0348, +0.1067] \\
GR00T / Open cabinet & \textbf{0.6370} & 0.6425 & 1.1156 & \textbf{0.8748} & 0.4786 & \textbf{0.2322} & [+0.1232, +0.3845] \\
$\pi_{0.5}$ / Insert battery & 0.8263 & \textbf{0.8250} & \textbf{0.8734} & 0.8783 & \textbf{0.0470} & 0.0533 & [-0.0157, +0.0035] \\
$\pi_{0.5}$ / Wipe bowl & \textbf{0.7212} & 0.7277 & \textbf{0.7359} & 0.7499 & \textbf{0.0146} & 0.0222 & [-0.0127, -0.0024] \\
$\pi_{0.5}$ / Reinsert USB & \textbf{0.6094} & 0.6164 & 0.6735 & \textbf{0.6606} & 0.0641 & \textbf{0.0442} & [-0.00005, +0.04273] \\
$\pi_{0.5}$ / Open cabinet & 0.5428 & \textbf{0.5352} & 0.5500 & \textbf{0.5380} & 0.0072 & \textbf{0.0027} & [-0.0020, +0.0104] \\
\bottomrule
\end{tabular*}
\end{table}